\pdfoutput=1
\documentclass[11pt]{article}

\usepackage[final]{acl}

\usepackage{times}
\usepackage{latexsym}

\usepackage[T1]{fontenc}
\usepackage[utf8]{inputenc}

\usepackage{booktabs}
\usepackage{tabularx}

\usepackage{microtype}

\usepackage{inconsolata}
\usepackage{multirow}

\usepackage[most]{tcolorbox}

\usepackage{makecell}
\usepackage{graphicx}
\usepackage[table,xcdraw]{xcolor}
\newif\ifcomments
\commentstrue  % Comment out this line to hide all comments.

\title{WISE: Weak-Supervision-Guided Step-by-Step Explanations for Multimodal LLMs in Image Classification}

\author{
\textbf{Yiwen Jiang} $^{1,2}$ \quad \textbf{Deval Mehta} $^{1,2,\dagger}$ \quad \textbf{Siyuan Yan} $^{1,2}$ \quad \textbf{Yaling Shen} $^{2}$ \\ \textbf{Zimu Wang} $^{2}$ \quad \textbf{Zongyuan Ge} $^{2,\dagger}$
\\
\textsuperscript{1}Faculty of Engineering, Monash University, Melbourne, Australia \\
\textsuperscript{2}AIM for Health Lab, Faculty of IT, Monash University, Melbourne, Australia \\
\texttt{\{yiwen.jiang, deval.mehta, zongyuan.ge\}@monash.edu}
}

\usepackage{amsmath}

\begin{document}
\maketitle

\renewcommand{\thefootnote}{\fnsymbol{footnote}}
\footnotetext[2]{Corresponding authors.}
\renewcommand{\thefootnote}{\arabic{footnote}}
\footnotetext[1]{Data and codes are available on: 
\href{https://github.com/yiwenJG/WISE-MCoT}{\url{https://github.com/yiwenJG/WISE-MCoT}}}

\begin{abstract}
Multimodal Large Language Models (MLLMs) have shown promise in visual-textual reasoning, with Multimodal Chain-of-Thought (MCoT) prompting significantly enhancing interpretability. However, existing MCoT methods rely on rationale-rich datasets and largely focus on inter-object reasoning, overlooking the intra-object understanding crucial for image classification. To address this gap, we propose WISE, a \underline{W}eak-superv\underline{I}sion-guided \underline{S}tep-by-step \underline{E}xplanation method that augments any image classification dataset with MCoTs by reformulating the concept-based representations from Concept Bottleneck Models (CBMs) into concise, interpretable reasoning chains under weak supervision. Experiments across ten datasets show that our generated MCoTs not only improve interpretability by 37\% but also lead to gains in classification accuracy when used to fine-tune MLLMs\footnotemark[1]. Our work bridges concept-based interpretability and generative MCoT reasoning, providing a generalizable framework for enhancing MLLMs in fine-grained visual understanding.
\end{abstract}

\section{Introduction}

Deep Learning (DL) models have achieved remarkable performance, powering applications in various domains. However, DL architectures are inherently "black-box" which often result in limited interpretability of the underlying decision-making processes \citep{papernot2017practical}. Thus, an increasing amount of attention has been directed toward developing DL models that are either inherently interpretable, or capable of producing explicit chains of reasoning that reveal their conclusions.

In the realm of generative DL, very recent Multimodal Large Language Models (MLLMs) have become a powerful paradigm for joint visual-textual processing and importantly introducing reasoning \citep{liu2023visual} to decision-making. To enhance their interpretability, recent work has proposed Multimodal Chain-of-Thought (MCoT) reasoning \citep{zhang2023multimodal}, which simulates human-like step-by-step inference by breaking down complex problems into sequential, interpretable reasoning steps. This has proved to improve interpretability via reasoning, which ultimately enhances the performance on multimodal tasks \citep{chen-etal-2024-m3cot}.

\begin{figure}[t]
  \includegraphics[width=\columnwidth]{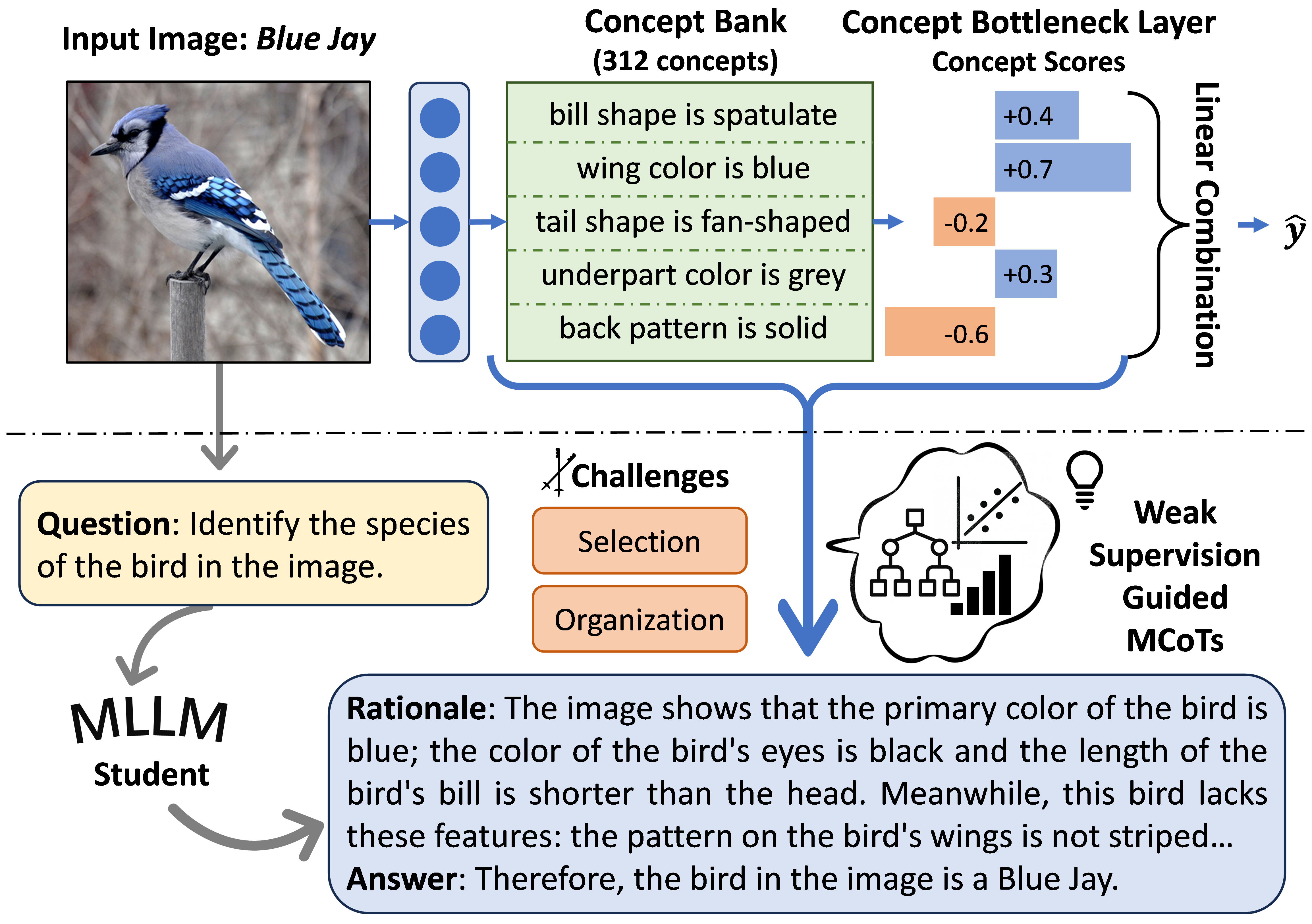}
  \caption{Exhaustive concept sets and the intrinsic linear combination in CBMs hinder their direct transformation into CoT. Our method addresses these challenges through a weak-supervision-guided reformulation of the bottleneck layer into concise textual rationales.}
  \label{fig:cbm2mllms}
\end{figure}

Datasets with rationales are essential for eliciting MCoT reasoning in MLLMs. However, they are typically constructed through costly human annotations or automatic generation via prompting LLMs, both of which pose challenges in ensuring high data quality. Existing MCoT datasets, such as CoMT \citep{cheng2025comt}, primarily target complex inter-object reasoning, and none to date focus on image classification tasks involving intra-object understanding. However, prior work \citep{VLMClassifier} has shown that integrating classification-focused data into MLLMs training, even without MCoT supervision, can enhance higher-level visual capabilities such as visual question answering and reasoning, highlighting classification as a foundation for multimodal reasoning. Since many MCoT approaches \citep{chen2023see} begin by detecting objects in the image, classification naturally contributes to reasoning accuracy. However, the problem of generating high-quality natural language MCoT tailored to image classification through automated methods remains largely unexplored.

Parallel to recent efforts in generative DL, research on discriminative models has predominantly relied on concept-based analysis \citep{kim2018interpretability}, exemplified by Concept Bottleneck Models (CBMs) \citep{koh2020concept}, to enhance interpretability \citep{mehta2025interpretable}. CBMs aim to associate each neuron with a human-understandable concept. In image classification, they map visual representations to a set of textual concepts, from which predictions are derived via a linear combination of concept scores. CBMs also offer intervenability: Within a human-in-the-loop framework \citep{yan2023towards}, humans can alter predictions by adjusting wrongly activated concepts, enabling direct control over model behavior \citep{koh2020concept}. Recent CBM work has pushed the paradigm forward by enabling fully automated language grounding (\citealp{oikarinen2023label}; \citealp{yang2023language}; \citealp{yan2023learning}), which prompts pre-trained Large Language Models (LLMs) with category names to generate candidate concepts, selects representative ones to construct a concept bank, and employs multimodal models such as CLIP \citep{radford2021learning} to align images and concepts via image-text scoring, forming a Concept Bottleneck Layer (CBL).

Inspired by the success of CBMs in interpretable image classification, we pose a natural question: \textbf{Can the CBL be reformulated as MCoT to facilitate the training of MLLMs?} A naive approach that directly transforms concepts into natural language is not feasible due to the extensive set of possible concepts and the importance of their ordering (Figure~\ref{fig:cbm2mllms}). Addressing this question requires overcoming two key challenges: (1) Selecting appropriate concepts to serve as components of the MCoT. Unlike CBMs, which score and combine all concepts during inference, generative models cannot feasibly incorporate such exhaustive representations. For instance, the CUB dataset \citep{wah2011caltech} contains 312 annotated bird attributes, making it impractical to reflect all CBM neurons in a single rationale; (2) Organizing the selected concepts into coherent reasoning chains that align with human cognitive patterns. In CBMs, each concept may contribute positively or negatively to a prediction, supporting or refuting specific categories. Moreover, concepts vary in their contribution to the final decision. These aspects must be carefully modeled to construct effective  MCoTs.

Motivated by the principle of weak-to-strong generalization \citep{burns2023weak}, we propose a novel \underline{W}eak-superv\underline{I}sion-guided \underline{S}tep-by-step \underline{E}xplanation method (\textbf{WISE}) for automatic MCoT generation. It reformulates the CBL as concept-driven natural language reasoning. Specifically, we use CLIP to score images against a concept bank and utilize CBMs for concept annotation. Leveraging the prior distribution between categories and concepts, we apply decision tree \citep{breiman1984cart} to construct Prior Trees. To capture instance-level variation and reflect the human tendency to organize concepts sequentially, we further design two instance-specific trees: an Affirmation Tree and an Elimination Tree. These trees are then combined and transformed into MCoTs. Finally, we fine-tune MLLMs using a curriculum learning strategy. Experiments across ten image classification datasets show that our method improves the interpretability of MLLMs by 37\% and enhances classification accuracy, despite being guided by models that are more lightweight than billion-parameter MLLMs.

Furthermore, our generated MCoTs align with human reasoning patterns and directly address the two previously identified challenges: by incorporating both discriminativeness and visual salience, they focus on a small set of critical concepts for decision-making; they also capture concept typicality and reflect both affirmative and counterfactual reasoning strategies. Overall, our main contributions are as follows:

$\bullet$ To the best of our knowledge, we are the first to bridge the previously separate paradigms of CBMs and MCoTs by proposing WISE, a weak-supervision-guided method that reformulates CBM representations into natural language MCoTs.

$\bullet$ Our method transforms any image classification dataset with category labels into an MCoT-augmented version, producing rationales that reflect human reasoning by integrating category typicality, instance-level distinctiveness, and both supportive and counterfactual evidence.

$\bullet$ We conduct experiments on ten image classification datasets, showing that our generated MCoTs improve the interpretability of MLLMs by 37\% while also enhancing classification accuracy.

\section{Related Work}

\noindent \textbf{Concept Bottleneck Models.} CBMs \citep{koh2020concept} are a prominent approach for designing inherently interpretable DL models, as detailed by \citet{zhou2018interpretable} and \citet{losch2019interpretability}. CBMs incorporate a concept bottleneck layer preceding the final fully connected layer, where each neuron represents a human-interpretable concept. \citet{yuksekgonul2023posthoc} and \citet{oikarinen2023label} proposed data-efficient methods to convert any DL models into CBMs without training from scratch. CLIP-based CBMs (\citealp{jiang-etal-2025-enhancing}; \citealp{shang2024incremental}; \citealp{oikarinen2023label}; \citealp{yang2023language}; \citealp{yan2023learning}) have leveraged vision-language alignment learned during pre-training \citep{radford2021learning} to eliminate the need for concept-level manual annotation, enabling automatic concept bank construction via LLMs and achieving competitive performance with "black-box" models. Recent studies such as Concept Agent \citep{jiang-etal-2025-enhancing} and LM4CV \citep{yan2023learning} explores concise concept banks to reduce redundancy, but the intrinsic CBM architecture, the linear combination of individual concept scores, limits their direct use in CoT reasoning for MLLMs. Crucially, our method dynamically selects concepts for reasoning on a per-image basis, instead of relying on a predefined, fixed set.

\noindent \textbf{MCoT Reasoning for MLLM.} Multimodal Chain-of-Thought (MCoT) extends the Chain-of-Thought (CoT) paradigm \citep{wei2022chain} to MLLMs, aiming to enhance their ability to perform stepwise reasoning across diverse input modalities. CoT improves both transparency and accuracy by decomposing complex problems into intermediate steps \citep{zhang2023multimodal}. In explainable image classification, MCoT incorporates visual inputs while generating rationales in natural language. To better structure the reasoning process, various topologies such as trees \citep{yao2023tree} and graphs \citep{besta2024graph} have been explored, enabling richer semantic composition and flexible backtracking. Prompt-based methods enable MLLMs to produce rationales at inference time via crafted instructions or in-context examples, requiring no additional training (\citealp{luo2024pkrd}; \citealp{zheng2024thinking}; \citealp{gao2024cantor}). In contrast, learning-based methods \citep{zhang2023multimodal} fine-tune MLLMs on annotated rationale data, making them more effective at implicitly acquiring reasoning patterns. Although several MCoT datasets are available \citep{chen2023see}, they primarily target reasoning over inter-object relations or rudimentary knowledge. To date, there remains a lack of rationale-annotated datasets tailored to image classification that emphasize intrinsic properties of individual objects.

\noindent \textbf{Weak-to-Strong Interpretability.} Weak-to-strong generalization \citep{burns2023weak} is a paradigm for eliciting the latent capabilities of powerful models through supervision provided by weaker models. Building on this idea, we explore whether weak supervision, including CBM \citep{koh2020concept}, CLIP \citep{radford2021learning}, Decision Tree \citep{breiman1984cart}, Linear Regression \citep{hastie2009elements}, and Bayesian Learning \citep{bayes1958essay}, can elicit interpretable MCoT reasoning from MLLMs. Among these, decision trees play a central role, recursively splitting the feature space based on input attributes to form a tree-like structure \citep{costa2023recent} where internal nodes represent decision rules and leaf nodes correspond to outcomes. This transparent structure enables intuitive interpretation of the decision-making process and has been extensively studied for applications in critical domains such as healthcare \citep{jiang2022medical}.

\section{Methodology}

\begin{figure*}[t]
  \centering
  \includegraphics[width=0.91\textwidth]{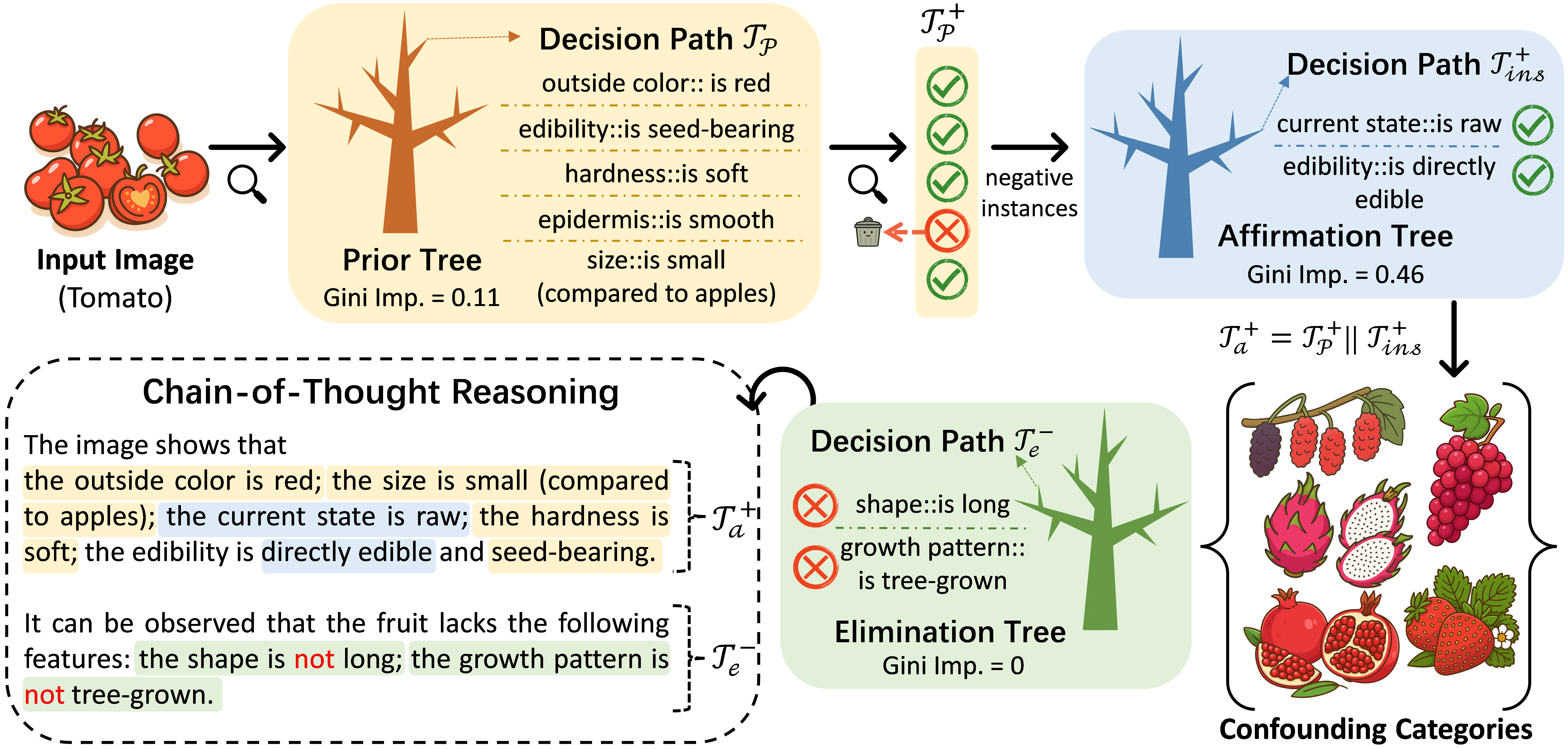}
  \caption{Overall Workflow of MCoT Generation by WISE. Given an LAD-F image \citep{zhao2019large}, WISE queries its \textit{Prior Tree} (Sec.~\ref{sec::prior}) to locate the decision path $\mathcal{T}_{p}$ and identify present concepts $\mathcal{T}_{p}^{+}$. It then builds the \textit{Affirmation Tree} (Sec.~\ref{sec::instance_tree}) from matching negatives. If these concepts $\mathcal{T}_{a}^{+}$ remain insufficient for class determination, WISE builds an \textit{Elimination Tree} $\mathcal{T}_{e}^{-}$ (Sec.~\ref{sec::instance_tree}) to refute remaining negatives until zero Gini impurity. The final concepts are sequentially verbalized as MCoT ($\mathcal{T}_{\text{MCoT}} = \mathcal{T}_{a}^+ \ \Vert \ \mathcal{T}_{e}^-$) using a template-based approach (Sec.~\ref{sec::rational}).}
  \label{fig:method}
\end{figure*}

\subsection{Problem Formulation}

Given a dataset $\mathcal{D} = {(x_{i}, y_{i})}_{i=1}^{\mathcal{K}}$, where $x_i \in \mathcal{X}$ denotes an image and $y_i \in \mathcal{Y}$ is one of $\mathcal{N}$ predefined categories, and a concept set $\mathcal{C} = {c_1, c_2, \dots, c_M}$ provided by humans or LLMs as the semantic basis for interpretation, the objective is to automatically generate textual rationales $R$ to guide the fine-tuning of MLLMs on $\mathcal{D}$.

We formulate this as modeling the joint distribution $P(\mathcal{Y}, R \mid \mathcal{X}, Q)$, where $Q$ denotes a task-specific prompt (e.g., \textit{Identify the species of the bird in the image}). The rationales $R$ are MCoTs constrained to reflect a subset of the concept set, i.e., $\text{Info}(R) \subseteq \mathcal{C}$, with $|\text{Info}(R)| \ll M$, enabling concise and faithful concept-driven reasoning. $\mathcal{Y}$, $R$ and $Q$ are represented as a sequence of language tokens in MLLMs.

\subsection{Concept Scoring for Visual Salience}

Humans tend to prioritize visually salient features as key cues for inference when recognizing objects. Motivated by this observation, we initiate rationale generation using Visual-Language Models (VLMs), such as CLIP \citep{radford2021learning}. CLIP consists of two encoders: an image encoder $\mathcal{I}:\mathcal{X} \to \mathcal{R}^d$ and a text encoder $\mathcal{T}:\mathcal{C} \to \mathcal{R}^d$, which project images and textual concepts into a shared $d$-dimensional embedding space. The concept score between an input image $x_i$ and a concept $c_m$ is calculated using cosine similarity:
\begin{equation}
    \label{eq:clip_concept_score}
    s(x_i, c_m) = \mathcal{I}(x_i) \cdot \mathcal{T}(c_m)
\end{equation}
This score quantifies the degree of cross-modal alignment and reflects the visual salience of concept $c_m$ in image $x_i$. To obtain concept-class alignment weights, we treat the concept score vector $s(x_i) \in {R}^M$ as a concept bottleneck layer and apply a softmax classifier trained with label supervision. This allows us to learn a weight vector $\mathbf{w}_{y_n} \in R^M$ for each class $y_n$, where each dimension reflects the relative importance and polarity (positive or negative) of a concept for that class:
\begin{equation}
    \label{eq:lr_softmax}
    P(y_n \mid x_i) = \frac{\exp(\mathbf{w}_{y_n}^\top s(x_i) + b_{n})}{\sum_{j=1}^{\mathcal{N}} \exp(\mathbf{w}_{y_j}^\top s(x_i) + b_j)}
\end{equation}
Here, $\mathbf{w}_{y_n}$ and $b_n$ denote the weight vector and bias term for class $y_n$. Finally, for an image $x_i$ labeled as $y_n$, we derive the binary annotation for each concept $c_m$ based on the sign of its contribution to the predicted class. Let $z_{i,m} \in \{0, 1\}$ denote the annotation of concept $c_m$ in $x_i$:
\begin{equation}
    \label{eq:concept_annotation}
        z_{i,m} =
  \begin{cases} 
    1, & \text{if } s(x_i, c_m) \cdot \mathbf{w}_{y_n}^{c_m} > 0 \\ 
    0, & \text{if } s(x_i, c_m) \cdot \mathbf{w}_{y_n}^{c_m} \leq 0
  \end{cases}
\end{equation}
For datasets which possess the ground-truth concept labels, the binary annotation step is not required. We then apply logistic regression, supervised by these labels, to map concept scores to probabilities, denoted as $P_{i,m}$ and determine a per-concept threshold optimized for macro F1 score. Concept instances with probabilities below the threshold, even if they are true positives, are relabeled as negative. This refinement retains only the most visually salient concepts.

\subsection{Category Typicality Tree Modeling}
\label{sec::prior}

During object recognition, humans often rely on stereotypical impressions associated with imagined categories. From a Bayesian perspective, this corresponds to a prior probability \citep{bayes1958essay}, specifically the category-to-concept prior denoted as $P(c_m \mid y_n)$ in our setting. To model this prior knowledge, we compute the prior distribution for each category–concept pair as follows:
\begin{equation}
    \label{eq:prior}
    P(c_m \mid y_n) = \frac{1}{|\mathcal{D}_{y_n}|} \sum_{i \in \mathcal{D}_{y_n}} P_{i,m} 
\end{equation}
Here, $\mathcal{D}_{y_n} = \{i \mid y_i = y_n\}$ denotes the set of training instances labeled with class $y_n$. This computation yields a prior matrix $\mathbf{P} \in \mathcal{R}^{N \times M}$, where each entry $[\mathbf{P}]_{n,m}$ represents the prior strength linking class $y_n$ and concept $c_m$. We use this matrix to build a category-specific decision tree for each $y_n$.

\textbf{Prior Tree.} The objective of the tree modeling is to discover the shortest decision path that distinguishes $y_n$ from the remaining classes, which are treated as negative samples. To this end, we identify the most salient concepts for $y_n$ by filtering the concept dimensions with $[\mathbf{P}]_{n,m} > 0.5$. These selected concepts serve as input features for the decision tree algorithm, which recursively select the concept that yields the highest information gain according to Gini impurity \citep{breiman1984cart} at each node. The resulting decision path for class $y_n$ is denoted as:
\begin{equation}
    \mathcal{T}_{p}(y_n) = \{c_{1}, c_{2}, \dots, c_{{p}}\}
\end{equation}
where $\mathcal{T}_{p}(y_n)$ is an ordered sequence of concepts forming the decision path, and $p$ denotes its length.

\subsection{Instance Distinctiveness Tree Modeling}
\label{sec::instance_tree}

While stereotypical impressions reflect the prototypical characteristics of a category, individual instances often exhibit distinctive features, i.e., deviations from the prototype. Such variations are essential for human reasoning. We capture this distinctiveness by also formulating a tree-based learning framework.  We posit that human reasoning often follows a two-step process to account for variations and organization of multiple concepts: (1) it begins with affirmation, supporting a hypothesis based on observed concepts, and (2) when evidence is insufficient, it proceeds to elimination, ruling out confounding options based on absent concepts. To reflect this strategy, we decompose our tree construction into two sequential stages.

\textbf{Affirmation Tree.} This tree builds upon the Prior Tree and focuses exclusively on positive concepts supporting the target class. For a given instance $(x_i, y_n)$, we begin by identifying the subset of prior concepts that are both part of the category-level decision path $\mathcal{T}_p(y_n)$ and observed in $x_i$. We denote this instance-specific subpath as:
\begin{equation}
    \mathcal{T}_p^+(x_i, y_n) = \{c \in \mathcal{T}_p(y_n) \mid z_{i,c} = 1\}
\end{equation}
We then retrieve a set of hard negative instances from $\mathcal{D}$—instances not labeled as $y_n$ but sharing the same set of activated prior concepts $\mathcal{T}_p^+(x_i, y_n)$. Since these instances cannot be distinguished from $x_i$ using the Prior Tree alone, additional instance-specific concepts are needed to further support.

To resolve this ambiguity, we extract additional concepts from $x_i$ that are not included in the prior decision path $\mathcal{T}_p(y_n)$ but are present in the instance. These distinctive, instance-specific concepts serve as input features for constructing the Affirmation Tree, and their resulting decision path is denoted as:
\begin{equation}
    \mathcal{T}_{\text{ins}}^+(x_i, y_n) = \{c_{1}, c_{2}, \dots, c_{\text{ins}^+}\}
\end{equation}
where $c_{j} \in \mathcal{C} \setminus \mathcal{T}_p(y_n)$ and $z_{i, c_{j}} = 1$. Finally, the Affirmation Tree path is defined as:
\begin{equation}
    \mathcal{T}_{a}^+(x_i, y_n) = \mathcal{T}_{p}^+(x_i, y_n) \ \Vert \ \mathcal{T}_{\text{ins}}^+(x_i, y_n)
\end{equation}
Here, \( \Vert \) denotes path concatenation. The resulting concepts within the path is reordered according to their prior probabilities that reflect relative importance. In cases where the leaf node of the Affirmation Tree $\mathcal{T}_{a}^+(x_i, y_n)$ retain non-zero Gini impurity, an Elimination Tree is constructed to further disambiguate confounding classes.

\textbf{Elimination Tree.} Let $y_c$ denote the set of confounding classes that causes non-zero Gini impurity at the leaf node of the Affirmation Tree for the input $x_i$. The goal of the Elimination Tree is to exclude these classes by leveraging concepts that are absent in $x_i$. Specifically, we collect negative instances as $\mathcal{D}_{y_c} = \{i \mid y_i \in y_c\} $ and identify true negative concepts in $x_i$ as input features for decision tree construction. The resulting decision path is defined as:
\begin{equation}
    \mathcal{T}_{e}^-(x_i, y_n) = \{c_{1}, c_{2}, \dots, c_{\text{ins}^-}\}
\end{equation}
where $c_{j} \in \mathcal{C}$ and $z_{i, c_{j}} = 0$. These concepts are absent in $x_i$ but frequently occur in the confounding classes $y_c$, thereby complementing the insufficient evidence provided by the Affirmation Tree. If non-zero Gini impurity persists after applying both trees, we recommend expanding the concept bank $\mathcal{C}$, indicating that the current set of concepts is insufficient to distinguish between categories.

\subsection{Tree-Guided Rationale for MLLMs}
\label{sec::rational}

We define the final MCoT decision path as the concatenation of the Affirmation Tree and the Elimination Tree for a given instance $(x_i, y_n)$:
\begin{equation}
    \mathcal{T}_{\text{MCoT}}(x_i, y_n) = \mathcal{T}_{a}^+(x_i, y_n) \ \Vert \ \mathcal{T}_{e}^-(x_i, y_n)
\end{equation}
The resulting chain captures both supportive and exclusionary reasoning steps, accounting for both category-level prototypicality and instance-level distinctiveness. To convert this structured path into natural language rationales, we design a template-based generation module that verbalizes each concept $c_k$ into a descriptive clause, conditioned on its semantics and polarity. As shown in Figure~\ref{fig:method}, the resulting clauses are then sequentially composed into a coherent explanation that mirrors the underlying reasoning logic. Future work may explore leveraging generative LLMs to polish and rephrase the rationales. In our experiments, we retain the template-based method to facilitate concept clause extraction and evaluation via regular expressions.

\subsection{Fine-tuning MLLMs with MCoTs}

Rather than immediately fine-tuning the MLLMs to perform concept-driven reasoning, we begin with task adaptation to guide the model in learning how to ground individual concepts in images. Following the principle of curriculum learning \citep{bengio2009curriculum}, we adopt a two-stage fine-tuning strategy that gradually increases task complexity.

In the first stage, we create a question-answering dataset by templating each annotated concept into a natural language QA pair (e.g., $Q$: \textit{What color are the bird's feathers?} $A$: \textit{Blue}), enabling the model to associate visual features with individual concepts. In the second stage, we further fine-tune the model on the Tree-guided MCoT dataset to enable compositional reasoning over multiple concepts.

\section{Experiments}

\begin{table*}[t]
\centering
\resizebox{0.80\textwidth}{!}{%
\begin{tabular}{cccccccccc}
\Xhline{1pt}
\multicolumn{2}{c}{\textbf{Model}}                                               & \multicolumn{4}{c}{\textbf{Weak Supervisors}}                                                                                                           & \multicolumn{4}{c}{\textbf{MLLM (Qwen2-VL)}}                                                                                   \\ \hline
\multicolumn{2}{c}{Dataset}                                                      & CLIP                      & CBM                           & DT                                     & NBC                                                & ZS-IO                     & ZS-MCoT                       & IT-IO                     & \textbf{IT-MCoT (ours)}                         \\ \Xhline{1pt}
                            & \multicolumn{1}{c|}{acc.}                          & 59.30                     & 45.94                         & 28.65                                  & \multicolumn{1}{c|}{60.30}                         & 33.21                     & 26.89                         & 82.40                     & {\color[HTML]{000000} 83.69 \textcolor{green!80!black}{(+1.29)}}   \\
\multirow{-2}{*}{CUB}       & \multicolumn{1}{c|}{\cellcolor[HTML]{EFEFEF}intp.} & \cellcolor[HTML]{EFEFEF}- & \cellcolor[HTML]{EFEFEF}55.62 & \cellcolor[HTML]{EFEFEF}\colorbox{cyan!50}{87.79} & \multicolumn{1}{c|}{\cellcolor[HTML]{EFEFEF}50.51} & \cellcolor[HTML]{EFEFEF}- & \cellcolor[HTML]{EFEFEF}55.20 & \cellcolor[HTML]{EFEFEF}- & \cellcolor[HTML]{EFEFEF}\colorbox{cyan!25}{65.03} \\
                            & \multicolumn{1}{c|}{acc.}                          & 64.71                     & 68.42                         & 56.97                                  & \multicolumn{1}{c|}{68.42}                         & 67.18                     & 57.89                         & 74.61                     & 74.61 (-)                              \\
\multirow{-2}{*}{SkinCon-3} & \multicolumn{1}{c|}{\cellcolor[HTML]{EFEFEF}intp.} & \cellcolor[HTML]{EFEFEF}- & \cellcolor[HTML]{EFEFEF}50.82 & \cellcolor[HTML]{EFEFEF}\colorbox{cyan!25}{79.92} & \multicolumn{1}{c|}{\cellcolor[HTML]{EFEFEF}56.64} & \cellcolor[HTML]{EFEFEF}- & \cellcolor[HTML]{EFEFEF}24.33 & \cellcolor[HTML]{EFEFEF}- & \cellcolor[HTML]{EFEFEF}\colorbox{cyan!50}{87.40} \\
                            & \multicolumn{1}{c|}{acc.}                          & 11.76                     & 60.06                         & 60.06                                  & \multicolumn{1}{c|}{60.06}                         & 22.60                     & 25.70                         & 60.37                     & 62.23 \textcolor{green!80!black}{(+1.86)}                          \\
\multirow{-2}{*}{SkinCon-9} & \multicolumn{1}{c|}{\cellcolor[HTML]{EFEFEF}intp.} & \cellcolor[HTML]{EFEFEF}- & \cellcolor[HTML]{EFEFEF}50.15 & \cellcolor[HTML]{EFEFEF}\colorbox{cyan!25}{91.67} & \multicolumn{1}{c|}{\cellcolor[HTML]{EFEFEF}53.55} & \cellcolor[HTML]{EFEFEF}- & \cellcolor[HTML]{EFEFEF}28.76 & \cellcolor[HTML]{EFEFEF}- & \cellcolor[HTML]{EFEFEF}\colorbox{cyan!50}{93.45} \\
                            & \multicolumn{1}{c|}{acc.}                          & 96.83                     & 73.36                         & 57.41                                  & \multicolumn{1}{c|}{86.63}                         & 79.67                     & 82.49                         & 97.50                     & 97.62 \textcolor{green!80!black}{(+0.12)}                          \\
\multirow{-2}{*}{LAD-A}     & \multicolumn{1}{c|}{\cellcolor[HTML]{EFEFEF}intp.} & \cellcolor[HTML]{EFEFEF}- & \cellcolor[HTML]{EFEFEF}60.18 & \cellcolor[HTML]{EFEFEF}\colorbox{cyan!50}{82.89} & \multicolumn{1}{c|}{\cellcolor[HTML]{EFEFEF}50.45} & \cellcolor[HTML]{EFEFEF}- & \cellcolor[HTML]{EFEFEF}55.90 & \cellcolor[HTML]{EFEFEF}- & \cellcolor[HTML]{EFEFEF}\colorbox{cyan!25}{82.78} \\
                            & \multicolumn{1}{c|}{acc.}                          & 75.56                     & 51.94                         & 44.08                                  & \multicolumn{1}{c|}{76.84}                         & 63.30                     & 67.53                         & 94.55                     & 94.70 \textcolor{green!80!black}{(+0.15)}                          \\
\multirow{-2}{*}{LAD-E}     & \multicolumn{1}{c|}{\cellcolor[HTML]{EFEFEF}intp.} & \cellcolor[HTML]{EFEFEF}- & \cellcolor[HTML]{EFEFEF}55.92 & \cellcolor[HTML]{EFEFEF}\colorbox{cyan!25}{82.12} & \multicolumn{1}{c|}{\cellcolor[HTML]{EFEFEF}51.49} & \cellcolor[HTML]{EFEFEF}- & \cellcolor[HTML]{EFEFEF}69.44 & \cellcolor[HTML]{EFEFEF}- & \cellcolor[HTML]{EFEFEF}\colorbox{cyan!50}{98.74} \\
                            & \multicolumn{1}{c|}{acc.}                          & 32.49                     & 13.88                         & 9.75                                   & \multicolumn{1}{c|}{19.04}                         & 21.17                     & 21.20                         & 59.02                     & 61.61 \textcolor{green!80!black}{(+2.59)}                          \\
\multirow{-2}{*}{LAD-H}     & \multicolumn{1}{c|}{\cellcolor[HTML]{EFEFEF}intp.} & \cellcolor[HTML]{EFEFEF}- & \cellcolor[HTML]{EFEFEF}57.50 & \cellcolor[HTML]{EFEFEF}\colorbox{cyan!25}{60.38} & \multicolumn{1}{c|}{\cellcolor[HTML]{EFEFEF}51.52} & \cellcolor[HTML]{EFEFEF}- & \cellcolor[HTML]{EFEFEF}23.28 & \cellcolor[HTML]{EFEFEF}- & \cellcolor[HTML]{EFEFEF}\colorbox{cyan!50}{83.44} \\
                            & \multicolumn{1}{c|}{acc.}                          & 73.33                     & 43.71                         & 36.06                                  & \multicolumn{1}{c|}{61.64}                         & 70.14                     & 70.05                         & 93.47                     & 93.17 \textcolor{red!60!white}{(-0.30)}                          \\
\multirow{-2}{*}{LAD-F}     & \multicolumn{1}{c|}{\cellcolor[HTML]{EFEFEF}intp.} & \cellcolor[HTML]{EFEFEF}- & \cellcolor[HTML]{EFEFEF}56.23 & \cellcolor[HTML]{EFEFEF}\colorbox{cyan!25}{76.77} & \multicolumn{1}{c|}{\cellcolor[HTML]{EFEFEF}50.65} & \cellcolor[HTML]{EFEFEF}- & \cellcolor[HTML]{EFEFEF}74.41 & \cellcolor[HTML]{EFEFEF}- & \cellcolor[HTML]{EFEFEF}\colorbox{cyan!50}{96.67} \\
                            & \multicolumn{1}{c|}{acc.}                          & 71.68                     & 51.21                         & 42.34                                  & \multicolumn{1}{c|}{75.25}                         & 47.92                     & 48.30                         & 94.10                     & 93.88 \textcolor{red!60!white}{(-0.22)}                          \\
\multirow{-2}{*}{LAD-V}     & \multicolumn{1}{c|}{\cellcolor[HTML]{EFEFEF}intp.} & \cellcolor[HTML]{EFEFEF}- & \cellcolor[HTML]{EFEFEF}58.57 & \cellcolor[HTML]{EFEFEF}\colorbox{cyan!25}{75.61} & \multicolumn{1}{c|}{\cellcolor[HTML]{EFEFEF}52.04} & \cellcolor[HTML]{EFEFEF}- & \cellcolor[HTML]{EFEFEF}74.70 & \cellcolor[HTML]{EFEFEF}- & \cellcolor[HTML]{EFEFEF}\colorbox{cyan!50}{95.12} \\ \hline
                            & \multicolumn{1}{c|}{acc.}                          & 60.71                     & 51.07                         & 41.92                                  & \multicolumn{1}{c|}{63.52}                         & 50.65                     & 50.23                         & 82.00                     & \textbf{82.69} \textcolor{green!80!black}{(+0.69)}                          \\
\multirow{-2}{*}{Average}   & \multicolumn{1}{c|}{\cellcolor[HTML]{EFEFEF}intp.} & \cellcolor[HTML]{EFEFEF}- & \cellcolor[HTML]{EFEFEF}55.62 & \cellcolor[HTML]{EFEFEF}\colorbox{cyan!25}{79.64} & \multicolumn{1}{c|}{\cellcolor[HTML]{EFEFEF}52.11} & \cellcolor[HTML]{EFEFEF}- & \cellcolor[HTML]{EFEFEF}50.75 & \cellcolor[HTML]{EFEFEF}- & \cellcolor[HTML]{EFEFEF}\textbf{\colorbox{cyan!50}{87.83}} \\ \Xhline{1pt}
\end{tabular}%
}
\caption{\textbf{Main results} on 8 concept-annotated datasets, reporting \textit{classification accuracy} (acc.) and \textit{interpretability} (intp.) for weak supervisors and MLLM. \textit{CLIP} is used in a zero-shot setting. The remaining methods are CLIP-based weak supervisors: Concept Bottleneck Model (\textit{CBM}), Decision Tree (\textit{DT}), and Naive Bayes Classifier (\textit{NBC}). \textit{ZS-IO} and \textit{ZS-MCoT} denote zero-shot input-output QA and zero-shot MCoT. \textit{IT-IO} is instruction tuning without rationales; \textit{IT-MCoT (ours)} integrates MCoTs derived from weak supervision. Accuracy gains over the \textit{IT-IO} baseline are indicated in \textcolor{green!80!black}{green} (improvement) and \textcolor{red!60!white}{red} (decline). Top-1 and Top-2 interpretability scores are highlighted in blue.}
\label{tab:main-result-table}
\end{table*}

\subsection{Datasets}

We evaluate our method on ten fine-grained image classification datasets across various domains and scales, including CUB \citep{wah2011caltech}, SkinCon \citep{daneshjou2022skincon}, LAD \citep{zhao2019large}, Oxford-Flowers \citep{nilsback2008automated}, and Oxford-Pets \citep{parkhi2012cats}. Among these, CUB, SkinCon, and LAD offer image-level concept annotations that enable quantitative evaluation of MCoT interpretability.

CUB contains 200 bird species, with each image annotated using 312 binary concept labels.
SkinCon is a dermatology dataset with hierarchical skin disease labels. Following \citet{daneshjou2022skincon} and \citet{pang2024integrating}, we construct two variants: SkinCon (3-class), using three coarse-grained disease categories and 22 concepts that appear in at least 50 images; and SkinCon (9-class), covering nine fine-grained categories with all 48 concepts. LAD consists of five sub-datasets covering animals (LAD-A), electronics (LAD-E), hairstyles (LAD-H), fruits (LAD-F), and vehicles (LAD-V). It includes 230 categories and 359 concepts in total. Following \citet{jiang-etal-2025-enhancing}, we construct concept banks for Oxford-Flowers and Oxford-Pets by prompting GPT-4o \citep{openai2024gpt4ocard}.

\subsection{Experimental Details}

\textbf{Weak Supervisors.} We evaluate both accuracy and interpretability of four weak supervisors, which are used to guide MLLMs in generating MCoTs: (1) \textbf{CLIP-zero-shot} \citep{radford2021learning} Serves as a baseline to showcase the pretrained vision-language alignment and classification capabilities of CLIP. (2) \textbf{CLIP-based CBM} (\citealp{yuksekgonul2023posthoc}, \citealp{oikarinen2023label}, \citealp{yan2023learning}) Applies logistic regression on CLIP-derived concept scores. Interpretability is quantified by the polarity of the product between concept weights and scores, indicating each concept’s contribution on the prediction. (3) \textbf{CLIP-based Decision Tree} \citep{breiman1984cart} Constructs decision rules over CLIP-annotated concepts. As the tree lacks direct visual access, interpretability is measured by the correctness of CLIP concept labels along decision paths. When annotations are fully accurate, the explanation is exact. (4) \textbf{CLIP-based Naive Bayes Classifier} \citep{bayes1958essay} Models prior and conditional probabilities under a conditional independence assumption. Concept polarity is computed as the log-ratio between predicted and contrasting classes, indicating directional contribution.

\textbf{MLLM Baselines.} We evaluate the inherent image classification and reasoning capabilities of MLLMs under two settings. Due to the large label space, which makes it impractical to enumerate all candidate labels in the prompt, we adopt an open-set image classification setup \citep{VLMClassifier}, providing a more challenging and realistic evaluation than the closed-set setup used for weak supervisors. (1) \textbf{Zero-shot Input-Output} The model is directly prompted to identify the object in the image without any fine-tuning. (2) \textbf{Zero-shot MCoT} The model is prompted to reason step by step. To establish a fair and informative baseline, we augment the prompt “Let’s think step by step” \citep{kojima2022large} with a comprehensive set of diverse concepts, systematically summarized from the concept bank to guide the reasoning process. GPT-4o \citep{openai2024gpt4ocard} is used to quantitatively evaluate the interpretability of the reasoning. To isolate the effect of MCoTs, we further include a fine-tuning-based baseline: (3) \textbf{Instruction Tuning (w/o MCoTs)}, The model is fine-tuned on question-answer pairs without explanatory rationales.

\textbf{Implementation.} We conduct experiments using Qwen2-VL-7B-Instruct \citep{wang2024qwen2} as the target MLLM and adopt CLIP ViT-L/14 as the backbone \citep{radford2021learning} of weak supervisors for most datasets, with a parameter count that is approximately 5\% of the target model. For SkinCon, we use MAKE \citep{yan2025make}, pretrained on the million-scale dermatology dataset Derm1M \citep{yan2025derm1m}, as the backbone. We fine-tune the MLLM using LoRA \citep{hu2022lora} with rank 8 for 10 epochs, employing a total batch size of 16 and a learning rate of $1 \times 10^{-4}$. All experiments are conducted on 4 NVIDIA RTX A5000 GPUs.

\textbf{Evaluation Metrics.} For MLLM-based methods, we report classification accuracy and the interpretability. Interpretability is quantified as the proportion of concept polarities in the rationales that agree with expert-annotated binary labels, following the standard CBM evaluation protocol. Since our approach reasons over a subset of concepts, this measure corresponds to concept precision.

\subsection{Main Results}

Table~\ref{tab:main-result-table} presents the overall evaluation results of our method on eight concept-annotated datasets.

\textbf{Interpretability.} Weak supervisors effectively guide MLLMs to acquire concept-driven reasoning abilities for image classification, achieving an average interpretability score of 87.83\% across eight datasets, which is unexpectedly 8.2\% higher than the best-performing weak supervisor (decision trees). On only two datasets does the MLLM’s interpretability fall short of its supervisor, further underscoring the central role of decision trees. Compared to the MLLM’s inherent zero-shot MCoT, our approach improves interpretability by 37\%, consistently outperforming across all datasets.

\textbf{Accuracy.} Differing from most CLIP-based CBMs, which often improve interpretability with slight compromises in accuracy, our method achieves both. Compared to instruction tuning using input-output pairs without MCoTs, our method improves the average accuracy by 0.69\%, with only minor drops observed on two datasets. Notably, when the MLLM’s inherent reasoning ability is weak, such as on LAD-H and SkinCon-9, where the average interpretability is only 26\%, guided reasoning yields substantially larger improvements in classification performance, with gains of up to 2\%. These results highlight the critical role of multi-step reasoning in enhancing final decision-making.

\textbf{Interpretability–Accuracy Trade-Off.} A clear trade-off between accuracy and interpretability is observed among weak supervisors: models with higher classification accuracy often exhibit lower interpretability. In contrast, our method integrates the strengths of all weak supervisors when constructing MCoTs, enabling interpretability to generalize effectively from weak to strong as the MCoTs align with human reasoning patterns. However, when the concept-level accuracy of zero-shot reasoning drops to 50.75\%, hallucinations tend to arise and propagate to the final predictions, thereby causing an average performance drop of 0.4\%.

\textbf{Datasets w/o Concept Annotation.} We further evaluate WISE on Oxford-Pets (37 classes) and Oxford-Flowers (102 classes), two datasets without concept annotations. Without MCoTs, instruction tuning achieves accuracies of 94.19\% and 98.72\%, respectively. Incorporating WISE yields 93.87\% and 98.88\%, indicating comparable performance. Moreover, the case studies in Appendix~\ref{appedix:examples} show that WISE substantially enhances interpretability.

% \usepackage{graphicx}
\begin{table}[t]
\centering
\resizebox{1\columnwidth}{!}{%
\begin{tabular}{cccccc}
\Xhline{1pt}
\textbf{} & \textbf{Pos} & \textbf{Neg} & \textbf{InCoT} & \textbf{XCoT} & \textbf{Bank} \\ \Xhline{1pt}
CUB       & 62.34          & 69.92          & 7                & 301             & 312               \\
SkinCon-3 & 66.01          & 89.16          & 17               & 22              & 22                \\
SkinCon-9 & 75.00          & 94.41          & 23               & 48              & 48                \\
LAD-A     & 82.78          & -              & 3                & 110             & 123               \\
LAD-E     & 98.62          & 100.00         & 3                & 72              & 75                \\
LAD-H     & 86.86          & 73.81          & 4                & 22              & 22                \\
LAD-F     & 97.29          & 89.47          & 5                & 53              & 58                \\
LAD-V     & 95.15          & 94.73          & 3                & 76              & 81                \\ \hline
Average   & 83.01          & 87.36          & 8                & 88              & 93                \\ \Xhline{1pt}
\end{tabular}%
}
\caption{\textbf{Analysis of MCoTs}. \textit{Pos} and \textit{Neg} denote the precision of supportive and refutational concepts of the MLLM-generated MCoTs on the test set, respectively. \textit{InCoT} indicates the average number of concepts used per image for reasoning, \textit{XCoT} denotes the number of unique concepts used at least once across all images, and \textit{Bank} refers to the total number of expert-defined concepts provided for weak supervision.}
\label{tab:cpt_analysis}
\end{table}

\subsection{MCoT Analysis}
\label{sec:mcot-analysis}

Table~\ref{tab:cpt_analysis} reports the decoupled precision of positive and negative concepts within the MCoTs on the test dataset, representing supportive reasoning and refutational elimination, respectively. In addition, the table presents the number of concepts used in the MCoTs and compares them with the total number of predefined concepts in the bank, highlighting the efficiency and dynamic selectivity of the reasoning.

\textbf{Concept Contribution.} The model achieves an average precision of 83\% for positive concepts and 87\% for negative concepts, indicating a balanced adoption of both reasoning strategies. These results suggest that the MLLM integrates both supportive and refutational reasoning without exhibiting performance bias. In the relatively simple LAD-A, the decisions can be made solely through supportive reasoning, with no reliance on refutational cues.

\textbf{Concept Efficiency.} Our MCoTs require only 8 concepts on average to complete reasoning across all datasets, consistent with the human intuition that a small number of concepts is sufficient for object identification. In contrast, traditional CBMs score all 312 concepts in the concept bank to reach a decision, which clearly contradicts this intuition. Although recent efforts \citep{yan2023learning} have reduced this number to 32 on CUB, the inherent scoring mechanism of CBM still leads to redundancy. Notably, our method achieves accurate classification of 200 bird species using only 7 concepts. This efficiency stems from the decision tree’s ability to capture logical dependencies among concepts. We also observe that the SkinCon dataset requires significantly more concepts. Manual inspection reveals that this is due to limitations of the concept bank: Some diseases share identical concept patterns, rendering them indistinguishable under the current representation. Future work should consider expanding the concept bank of SkinCon to improve its coverage and discriminative capacity.

\textbf{Dynamic Concept Selection.} We observe that although reasoning over a single image involves only 8 concepts on average, the MCoTs generated by WISE dynamically select concepts, resulting in 95\% of the concept bank being used at least once. This demonstrates a clear contrast to prior work \citep{jiang-etal-2025-enhancing}, which relies on a fixed concept bank and scores each concept independently.

\subsection{Ablation Study}

Table~\ref{tab:ablation} compares the concept precision of MCoTs generated under different strategies on CUB.

\textbf{Salient Concept Selection.} Selecting visually salient concepts substantially improves MLLMs' reasoning precision. CLIP’s vision-language alignment scores not only compensate for weak supervisors’ limited visual perception, but also reduce MLLMs’ reliance on imperceptible cues.

\textbf{Concept Organization.} We create a variant by shuffling the order of concepts in the generated MCoTs, resulting in rationales that are unordered and unstructured. This randomness increases the learning difficulty for the model. In contrast, we introduce human-aligned reasoning biases, reducing such uncertainty and improving concept precision.

\textbf{Captioning as MCoT.} To differentiate concept-driven reasoning from simple image captioning, we train MLLMs using all visual concepts present in the image, arranged in a predefined order. Result indicates caption-style MCoTs fail to facilitate the acquisition of concept-level reasoning.

\textbf{Variant Tree Construction.} We design two tree variants: category-specific and instance-specific. The former builds one tree per category, ignoring individual variability, whereas the latter relies on instance-level feature saliency, neglecting cross-category regularities. Consequently, both methods suffer reduced precision due to their lack of shared structure and sensitivity to individual differences.

\section{Discussion}

We demonstrated the efficiency and dynamic nature of concept selection in Section~\ref{sec:mcot-analysis}. Beyond these observations, WISE offers two notable advantages:

\textbf{Stepwise Transparent Reasoning.} Guided by decision trees, WISE generates MCoTs by selecting the most discriminative concept at each step based on information gain, progressively isolating the target class from negatives. Each selection explicitly conveys three elements to enhance transparency: (1) the rationale for the choice, (2) the negative classes excluded by this step, and (3) the remaining negatives contributing to non-zero Gini impurity. This stepwise process naturally aligns with the autoregressive behavior of MLLMs.

\textbf{Quantified Sufficiency.} Reasoning continues until a leaf node is reached, where the sufficiency of the reasoning chain is evaluated. A zero Gini impurity indicates complete reasoning, whereas a non-zero impurity (when no further splits are possible) reveals gaps in the current concept bank.

Overall, WISE achieves more efficient and concise concept usage than CLIP-based CBMs such as LM4CV \citep{yan2023learning} and neural-symbolic reasoning methods like Deep Concept Reasoner (DCR) \citep{pmlr-v202-barbiero23a}, while offering distinctive advantages. Unlike DCR’s permutation-invariant logical rules, WISE adopts a human-aligned reasoning paradigm that preserves the uniqueness of the generated MCoTs.

\section{Future Work}

Beyond enabling interpretable image classification with MLLMs, our work lays the foundation for more efficient human-in-the-loop frameworks \citep{yan2023towards}. Concept-based models allow human intervention at the concept level, providing direct control over model behavior \citep{koh2020concept}. By generating the minimal set of concepts needed for explanation, our method significantly improves intervention efficiency. This property is particularly valuable in domains such as clinical image diagnosis, where rapid and precise feedback is critical.

In addition, we introduce a new class of instruction data designed to promote intra-object understanding during MLLM pretraining. This resource can support the creation of large-scale MCoT datasets for both domain-specific and general applications. While our current experiments focus on a single dataset, future work should explore dataset integration and scaling toward training foundation-level MLLMs \citep{VLMClassifier}. Finally, the released MCoT dataset provides a practical benchmark for diverse vision tasks, including hallucination evaluation and mitigation \citep{bai2025hallucination}, enabling systematic comparison of hallucination-reduction strategies and advancing reliable MLLM reasoning.

% \usepackage{graphicx}
\begin{table}[]
\centering
\resizebox{0.80\columnwidth}{!}{%
\begin{tabular}{lccc}
\Xhline{1pt}
\multicolumn{1}{c}{\textbf{Method}} & \textbf{Overall} & \textbf{Pos-C} & \textbf{Neg-C} \\ \Xhline{1pt}
\textbf{Ours}                       & \textbf{65.03}  & \textbf{62.34} & \textbf{69.92} \\
- w/o Salience                      & 32.73           & 32.65          & 48.39          \\
- w/o Order                         & 61.46           & 56.22          & 69.13          \\
Captioning                          & 49.68           & 49.68          & -              \\
Instance Tree                       & 56.13           & 52.91          & 65.99          \\
Category Tree                       & 62.21           & 56.84          & 67.74          \\ \Xhline{1pt}
\end{tabular}%
}
\caption{\textbf{Ablation studies} on the CUB test set, assessing the effects of different components and variants on the concept-level precision (interpretability) of MCoTs.}
\label{tab:ablation}
\end{table}

\section{Conclusion}

We propose WISE, a method that reformulates the concept bottleneck layer into natural-language MCoTs guided by weak supervision, aligning the model’s reasoning with human thought patterns. This method is broadly applicable to any image classification dataset with category labels. Experiments show that the generated MCoTs yield a 37\% improvement in the interpretability of MLLMs.

\section*{Limitations}

One main limitation of our approach is its dependence on concept banks generated by prompting LLMs for datasets lacking predefined annotations. The quality of the resulting MCoTs is tied to the coverage of these banks, and insufficient concept sets may reduce the model’s ability to distinguish visually similar categories. For datasets with concept annotations, the quality of these annotations likewise affects MCoT interpretability.

Moreover, although we reformulate image classification as a question answering task, most datasets are designed with a fixed and limited label space. This mismatch can introduce concept conflicts when integrating multiple MCoT-augmented datasets. For example, generated rationales may fail to eliminate distractor labels that are absent from one dataset but present in others. These issues highlight the need for comprehensive datasets that adequately represent the target domain.

\bibliography{custom}

\appendix
\onecolumn

\section{Additional Examples}
\label{appedix:examples}

\begin{tcolorbox}[
left=1mm, right=1mm, top=0mm, bottom=0mm, colback=white, colframe=white, boxrule=0pt]

\noindent
\begin{minipage}[c]{0.3\textwidth}
    \centering
    \includegraphics[width=\linewidth]{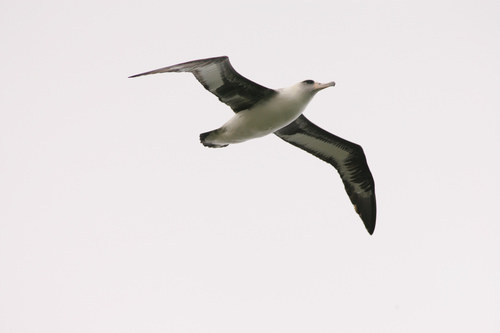}
\end{minipage}
\hfill
\begin{minipage}[c]{0.65\textwidth}
{\sffamily\small
\textbf{CUB.} The image shows that the color of the bird's breast is white; 
the color of the bird's underparts is white; the shape of the bird's bill 
is hooked (like a seabird's); the shape of the bird's wings is tapered; 
the shape of the bird is gull-like. It can be observed that the bird lacks 
the following features: the pattern on the bird's back is not solid; the 
color of the bird's under tail is not brown. Therefore, the bird in the image is laysan albatross.
}
\end{minipage}
%---------------------------------------------------------
\vspace{3mm}

\noindent
\begin{minipage}[c]{0.3\textwidth}
    \centering
    \includegraphics[width=\linewidth]{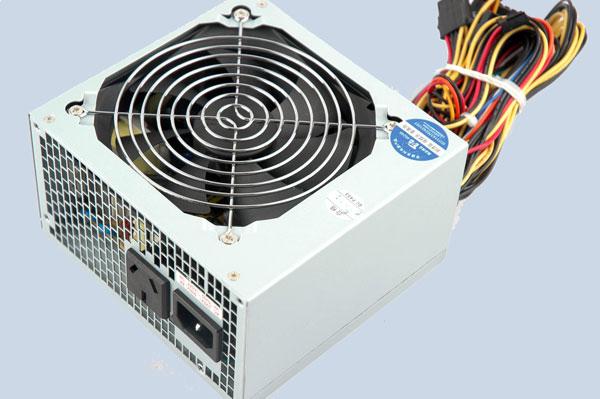}
\end{minipage}
\hfill
\begin{minipage}[c]{0.65\textwidth}
{\sffamily\small
\textbf{LAD-E.} The image shows that the size is big (compared to a mobile phone); the visible parts are a motor, a fan and indicator lights; the aim is display; the power rating is low-power. It can be observed that the device lacks the following features: the visible parts are not a plug. Therefore, the electronic device in the image is power supply unit.
}
\end{minipage}
%---------------------------------------------------------
\vspace{3mm}

\noindent
\begin{minipage}[c]{0.3\textwidth}
    \centering
    \includegraphics[width=\linewidth]{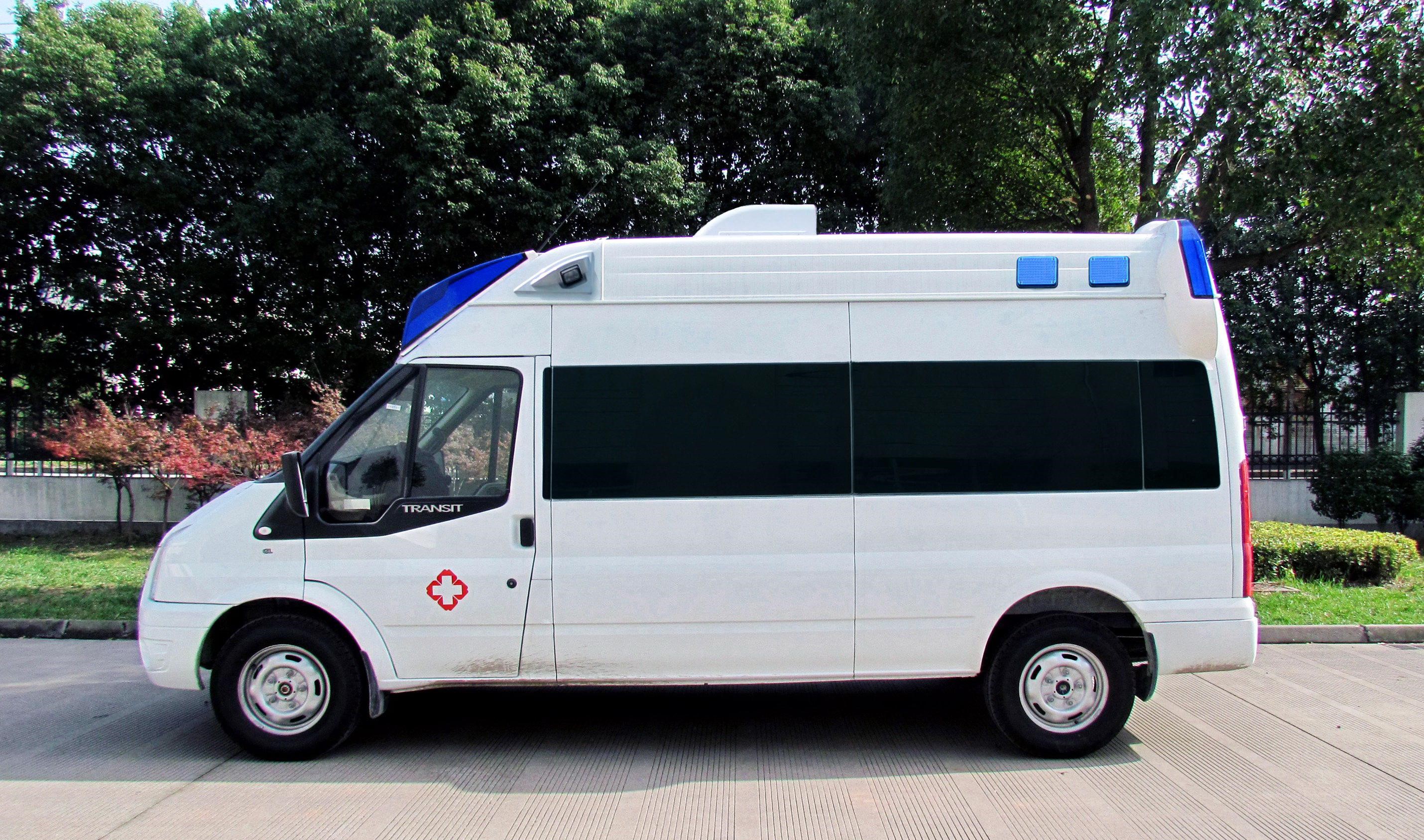}
\end{minipage}
\hfill
\begin{minipage}[c]{0.65\textwidth}
{\sffamily\small
\textbf{LAD-V.} The image shows that the color is white; the speed is fast-moving; the parts are a number plate; the aim is rescuing; the price is expensive. It can be observed that the vehicle lacks the following features: the aim is not engineering; the power source is not gasoline. Therefore, the vehicle in the image is ambulance.
}
\end{minipage}
%---------------------------------------------------------
\vspace{3mm}

\noindent
\begin{minipage}[c]{0.3\textwidth}
    \centering
    \includegraphics[width=0.75\linewidth]{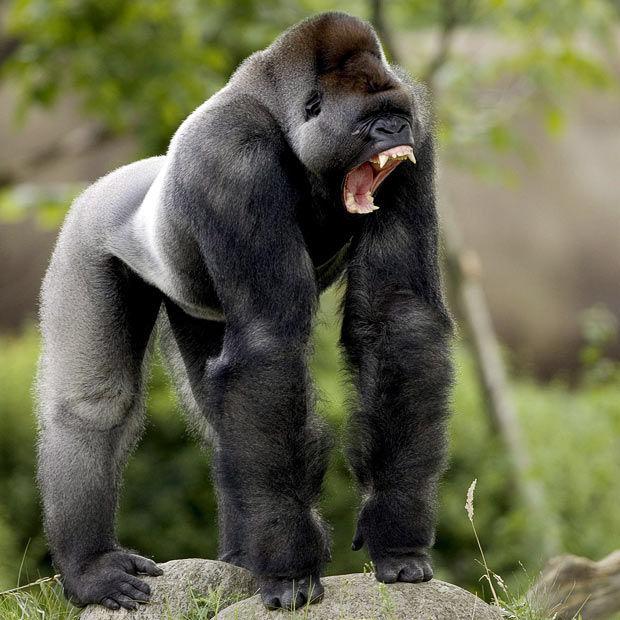}
\end{minipage}
\hfill
\begin{minipage}[c]{0.65\textwidth}
{\sffamily\small
\textbf{LAD-A.} The image shows that the color of the animal is black; the teeth are buck-toothed; the diet is leaf-based; the habitat is jungle. Therefore, the animal in the image is gorilla.
}
\end{minipage}
%---------------------------------------------------------
\vspace{3mm}

\noindent
\begin{minipage}[c]{0.3\textwidth}
    \centering
    \includegraphics[width=\linewidth]{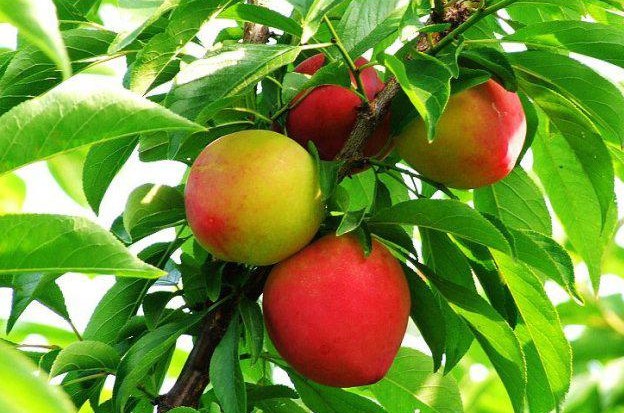}
\end{minipage}
\hfill
\begin{minipage}[c]{0.65\textwidth}
{\sffamily\small
\textbf{LAD-F.} The image shows that the outside color is red; the epidermis texture is peel-covered; the current state is complete; the hardness is soft; the edibility is common, directly edible and water-rich; the growth pattern is tree-grown; the medicinal property is mild. Therefore, the fruit in the image is plum.
}
\end{minipage}
%---------------------------------------------------------
\vspace{3mm}

\noindent
\begin{minipage}[c]{0.3\textwidth}
    \centering
    \includegraphics[width=\linewidth]{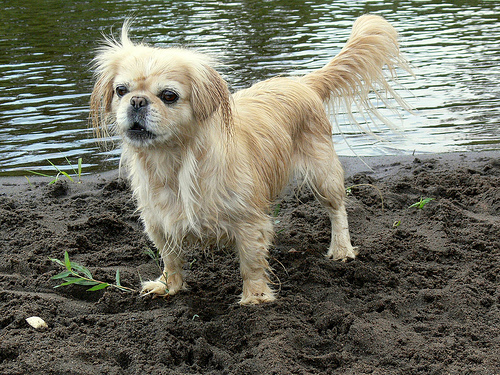}
\end{minipage}
\hfill
\begin{minipage}[c]{0.65\textwidth}
{\sffamily\small
\textbf{Oxford-Pets.} The image shows that the short, smooth coat with a pre-dominantly white color pattern; wide, deep muzzle with a well-defined stop; wide and prominent forehead with a gentle slope. Therefore, the pet in the image is Japanese Chin.
}
\end{minipage}
\end{tcolorbox}
\end{document}